%% file: main.tex
\pdfoutput=1
\documentclass[11pt]{article}

\usepackage{acl}

\usepackage{times}
\usepackage{latexsym}
\usepackage{color,soul}

\usepackage[T1]{fontenc}
\usepackage[utf8]{inputenc}

\usepackage{microtype}

\usepackage{algorithm}
\usepackage[algo2e]{algorithm2e} 
\usepackage{hyperref}
\usepackage{multirow}
\usepackage{booktabs}
\usepackage{amssymb}
\usepackage{adjustbox}
\usepackage[utf8]{inputenc}
\usepackage{enumitem}

\usepackage{diagbox}
	
\usepackage{xcolor, color, colortbl}
	
\definecolor{Gray}{gray}{0.95}
\definecolor{LightCyan}{rgb}{0.88,1,1}
\definecolor{celadon}{rgb}{0.67, 0.88, 0.69}
\newcolumntype{a}{>{\columncolor{LightCyan}}c}
\usepackage{amsmath}

\begin{document}

\title{Link, Synthesize, Retrieve: Universal Document Linking for Zero-Shot Information Retrieval}

\author{Dae Yon Hwang$^{1,2}$ \; Bilal Taha$^{2,3}$ \; Harshit Pande$^{1}$ \; Yaroslav Nechaev$^{1}$\\
$^1$ Amazon AGI\; $^2$ University of Toronto\; $^3$ Vector Institute\\
\texttt{daeyon.hwang@alumni.utoronto.ca}\;\; \texttt{bilal.taha@mail.utoronto.ca}\\
\texttt{pandeconscious@gmail.com}\;\; \texttt{\{dhwang, hppnd, nechaey\}@amazon.com}}

\maketitle

\begin{abstract}
Despite the recent advancements in information retrieval (IR), zero-shot IR remains a significant challenge, especially when dealing with new domains, languages, and newly-released use cases that lack historical query traffic from existing users. For such cases, it is common to use query augmentations followed by fine-tuning pre-trained models on the document data paired with synthetic queries. In this work, we propose a novel Universal Document Linking (UDL) algorithm, which links similar documents to enhance synthetic query generation across multiple datasets with different characteristics. UDL leverages entropy for the choice of similarity models and named entity recognition (NER) for the link decision of documents using similarity scores. Our empirical studies demonstrate the effectiveness and universality of the UDL across diverse datasets and IR models, surpassing state-of-the-art methods in zero-shot cases. The developed code for reproducibility is included in \url{https://github.com/eoduself/UDL}
\end{abstract}

%, and extending to question-answering (QA) paradigms \citet{chen2017reading}

\section{Introduction}

In information retrieval (IR), zero-shot learning is an essential problem that emerges when dealing with a new language or domain with little to no availability of the associated queries. Traditional IR methods primarily utilized sparse retrieval, while recent methods revolve around dense retrieval (DR), demonstrating the promising result \citep{neelakantan2022text}. Yet, using pre-trained DR directly on zero-shot cases results in substantial performance degradation, requiring dedicated fine-tuning \citep{izacard2021contriever, zhang-etal-2021-mr}.

One strategy for fine-tuning without relying on query traffic involves expanding the queries based on existing queries or documents with rule-based methods or language models (LMs) to obtain additional context in unseen domains \citep{wang-etal-2023-query2doc, jagerman2023query, weller-etal-2024-generative}. RM3~\citep{inproceedings} and AxiomaticQE~\citep{10.1007/978-3-030-15712-8_24} are classical ways to expand the queries with additional relevant terms while the recent studies indicate that large LMs (LLMs) can produce sophisticated synthetic data \citep{schick2021generating}, often resulting in better transfer learning than human-curated datasets \citep{liu2022wanli}. While LLMs like Gemini \citep{geminiteam2023gemini} generate superb synthetic queries for fine-tuning, devising a cost-effective way for IR remains challenging without additional recipes like dimensionality reduction \cite{hwang-etal-2023-embedtextnet}.

\begin{table}[!b]
\centering
\caption{Synthetic queries augmented by UDL.}
\scalebox{0.555}{
% \resizebox{8.4cm}{!}{%{2.4cm}{
\centering
\begin{tabular}{l|l|l} 
\toprule
\textbf{Document}                                                                                                                                                                                                                                                                                                                                                        & \begin{tabular}[c]{@{}l@{}}\textbf{Augmented } \\\textbf{query before UDL}\end{tabular}               & \begin{tabular}[c]{@{}l@{}}\textbf{Augmented }\\\textbf{query by UDL}\end{tabular}                        \\ 
\hline
\begin{tabular}[c]{@{}l@{}}In case of~allergic rhinitis,~you \\are still in group of subjects \\who can receive AstraZeneca's\\Covid-19 vaccine.\end{tabular}                                                                                                                                                                                                            & \begin{tabular}[c]{@{}l@{}}Subject of \\astrazeneca\\vaccination\end{tabular}                         & \multirow{2}{*}{\begin{tabular}[c]{@{}l@{}}\\Covid-19\\vaccination for\\allergic~rhinitis\end{tabular}}     \\ 
\cline{1-2}
\begin{tabular}[c]{@{}l@{}}With allergic rhinitis,~according\\to regulations of~the~Ministry of \\Health, you can still receive~\\the Covid-19 vaccine normally.\end{tabular}                                                                                                                                                                                            & \begin{tabular}[c]{@{}l@{}}Regulations of the \\Ministry of Health\\on allergic rhinitis\end{tabular} &                                                                                                           \\ 
\hline
\begin{tabular}[c]{@{}l@{}}Google Finance gives you\\free information.\end{tabular}                                                                                                                                                                                                                                                                                      & \begin{tabular}[c]{@{}l@{}}Google finance\\cost\end{tabular}                                          & \multirow{2}{*}{\begin{tabular}[c]{@{}l@{}}Which company \\gives the free\\quotes?\end{tabular}}          \\ 
\cline{1-2}
\begin{tabular}[c]{@{}l@{}}Sure, Yahoo Finance does\\this for FREE.\end{tabular}                                                                                                                                                                                                                                                                                         & \begin{tabular}[c]{@{}l@{}}Is yahoo finance\\free?\end{tabular}                                       &                                                                                                           \\ 
\hline
\begin{tabular}[c]{@{}l@{}}Most predict dire consequences\\if~GHGs continue to rise through\\the 21st century, which is what \\seems most likely.\end{tabular}                                                                                                                                                                                                           & \begin{tabular}[c]{@{}l@{}}Does GHG\\increase?\end{tabular}                                           & \multirow{2}{*}{\begin{tabular}[c]{@{}l@{}}\\What is the\\future of \\climate change?\end{tabular}}         \\ 
\cline{1-2}
\begin{tabular}[c]{@{}l@{}}There may be some tipping points\\that will accelerate climate change\\but we do not know~when each of \\these will become a problem.\end{tabular}                                                                                                                                                                                            & \begin{tabular}[c]{@{}l@{}}Acceleration of\\climate change\end{tabular}                               &                                                                                                           \\ 
\hline
\begin{tabular}[c]{@{}l@{}}Public health is a key issue– the state\\has a role in stopping people harming\\themselves – they may be harming\\themselves but the cost often falls on\\government through public healthcare,\\and therefore on all~taxpayers. Smoking\\also harms others through passive smoking.

\end{tabular}                                                                                                 & \begin{tabular}[c]{@{}l@{}}Why are we\\banning smoking?\end{tabular}                                  & \multirow{2}{*}{\begin{tabular}[c]{@{}l@{}}\\\\Do governments\\have the right to\\ban smokers?\end{tabular}}  \\ 
\cline{1-2}
\begin{tabular}[c]{@{}l@{}}Paternalistic Personal autonomy has to\\be the key to this debate. If people want\\to smoke – and the owner of the public\\place has no issue with that – it is not the\\role of the state to step in. All that is\\required is ensuring~that smokers are\\educated about the risks so that they can\\make an informed decision.\end{tabular} & \begin{tabular}[c]{@{}l@{}}Why the education\\needs for smoking\end{tabular}                          &                                                                                                           \\
\bottomrule
\end{tabular}
}
\label{tab:example_query}
\end{table}
To address the limitations of document-to-query generation, we propose a novel algorithm called Universal Document Linking (UDL), which offers an intuitive yet effective solution for zero-shot. This method links similar documents, aiding in the generation of synthetic queries spanning multiple documents. The UDL algorithm relies on selecting a similarity model based on term entropy and determining the link decisions using named entity recognition (NER) models. This approach facilitates the link decisions tailored to each dataset's unique characteristics, highlighting the universality of our method. Moreover, UDL is flexible to be combined with other query augmentations which reveals the high extensibility. With UDL, small LM can outperform LLM with a low cost. Table \ref{tab:example_query} presents examples demonstrating how UDL generates additional relevant queries that would not be generated by its absence. In this work, we make two main contributions: \textbf{(1)} Exploring the document linking for query augmentation with empirical studies which was not investigated previously, and \textbf{(2)} Introducing the UDL algorithm and demonstrating its effectiveness across diverse query augmentations, IR models, and datasets with varying tasks.

\section{Motivation}
\begin{figure}[!b]
\centering
\includegraphics[width=0.71\linewidth]{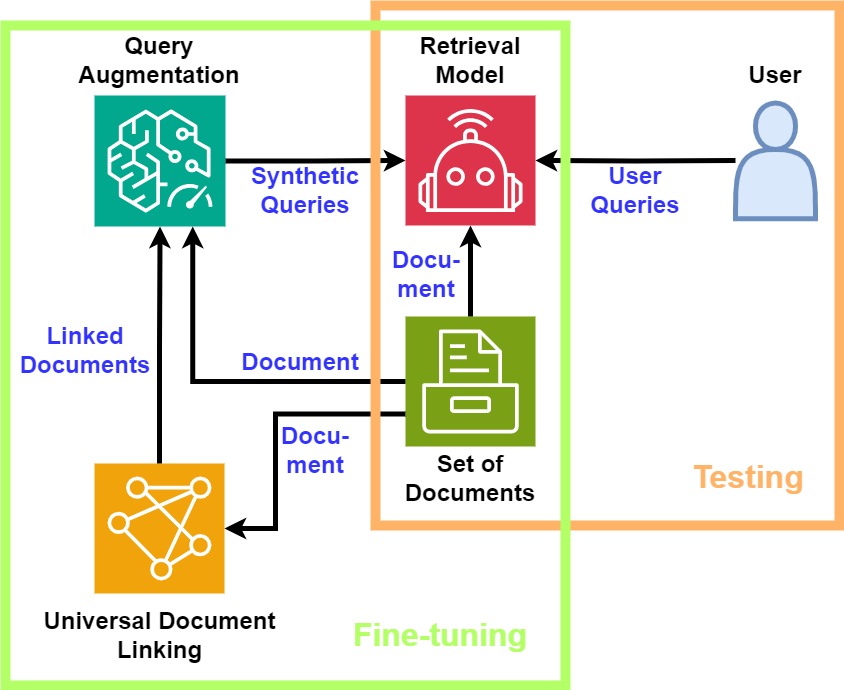} 
\caption{Overall zero-shot case. IR model is fine-tuned with synthetic queries, then interacted with user queries.}
\label{fig:overall_algorithm}
\end{figure}

Figure \ref{fig:overall_algorithm} illustrates the overall flow of fine-tuning a retrieval model in zero-shot scenario, where actual queries do not exist during fine-tuning. Instead, we use documents to generate synthetic queries, which aids the IR model in learning the distribution of the unseen domain~\citep{thakur2021beir}.

According to \citet{hwang-etal-2023-gan} and our initial findings (Table \ref{tab:size_query_appendix}), merely increasing the size of synthetic data doesn't consistently improve results. This is because query augmentation associates a synthetic query with a single document, whereas queries in datasets can be linked to multiple documents. Our insight from this led us to develop a method to link similar documents for the generation of synthetic queries that cover multiple documents.

% Note that TF-IDF is considered as lexical similarity which is based on the matching between word sets in documents and useful for identifying the unique feature (e.g. preventive of COVID) while pre-trained LM is employed as semantic similarity which relies on matching between the meaning of document and is helpful for contextual understanding

\begin{algorithm}[H]
 \KwData{A set of documents in each dataset}
\KwResult{Linked documents}
 \textbf{Parameters:} 
 Thresholds in similarity model $\gamma$ and score $\delta$, decision of similarity model $D_{M}$ and score $D_{T}$, pre-trained general NER $N_{g}$ and specialized NER $N_{s}$

\textbf{Step A.\;Decision of Similarity Model}\\[4pt]
 1.\;Measure TF-IDF in all documents \\[2pt]
 2.\;Calculate $Entropy$ for each term in TF-IDF across documents \\[2pt]
 3.\;\eIf{$D_{M}=\frac{\#\;of\;terms\;in\;Entropy\;>\;1}{\#\;of\;terms\;in\;Entropy\;\leq\;1} > \gamma$}
 {
    Use pre-trained LM as similarity model
 }
 {
    Use TF-IDF as similarity model
 }

 \textbf{Step B.\;Decision of Similarity Score}\\[4pt]
 1.\;\If{candidate documents not in English}{
  Translate to English
 }
 \;\>{2.}\;Eliminate the special characters in candidates \\[2pt]
\;\>{3.}\;$D_{T}$\;= 
\begin{equation}  % Use equation environment for display math
\begin{cases}
    \delta\;,\;\text{if } K_{N_{g}}\times\;V_{N_{s}} > K_{N_{s}}\times\;V_{N_{g}} \\
    1 - \delta\;,\;\text{otherwise}
\end{cases}
\end{equation}
$K$: Number of keywords from NER \\
$V$: Vocabulary size of NER

\textbf{Step C.\;Link Documents}\\[4pt]
1.\;Measure the cosine-similarity between candidate documents using a model from 
\textbf{A} \\[2pt]
2.\;\If{cosine-similarity > score from \textbf{B}}{
  Link documents
 }  

\caption{Universal Document Linking}
\label{algo:cl}
\end{algorithm}

\section{Universal Document Linking}

Algorithm \ref{algo:cl} outlines the procedural steps in the UDL. In the first step, denoted as \textbf{A}, the appropriate similarity model is selected for each dataset. We explore term frequency-inverse document frequency (TF-IDF) and pre-trained LM to derive document embeddings. Notably, TF-IDF considers lexical similarity, which is valuable for identifying unique features (e.g., disease like COVID), while pre-trained LM provides semantic similarity, aiding in contextual understanding. To determine the suitable similarity model, we initially compute TF-IDF scores for all documents, followed by calculating $D_{M}$ based on the Shannon entropy of terms using TF-IDF. Entropy values greater than 1 (i.e., numerator in $D_{M}$) describe high uncertainty since random variables have approximately uniform distribution in multiple classes. This concept is extended to the term entropy (Equation (\ref{eq:entropy})) where we calculate the entropy for each term across documents. 

% As shown in  ,
To accommodate the $D_{M}$ for the massive documents, we introduce the $\gamma$ value where articles and relatively common terms are mostly distributed in entropy greater 1 as expected (see Table \ref{tab:entropy-term}). Documents with an overwhelming presence of these terms are not desirable for TF-IDF since it can obscure the unique characteristics of documents, affecting link decisions. In such cases, considering semantically similar documents using pre-trained LM proves to be a more viable alternative.

% Entropy values greater than 1 (i.e., numerator in $D_{M}$) indicate high uncertainty, suggesting that terms are evenly distributed across most documents (e.g., definite articles. See Table \ref{tab:entropy-term}). Major terms exhibiting this situation (i.e., $D_{M}>\gamma\%$) are unsuitable for a TF-IDF similarity model, as it is less effective in revealing the unique characteristic of documents required for identifying similar documents for linking. In such cases, considering semantically similar documents using a pre-trained LM proves to be a more viable alternative.

After defining the similarity model, we proceed to determine the criteria in step \textbf{B} for deciding whether candidate documents should be linked. Each dataset contains varying levels of domain-specific terminology, which must be taken into account during document linking. To address this, we initially translated non-English documents into English using Google Translator \footnote{\scriptsize\url{https://github.com/ssut/py-googletrans}} to handle multilingual cases. After removing special characters, we compute $D_{T}$ based on the number of keywords extracted from NER models that are pre-trained on general ($N_{g}$) and specialized documents ($N_{s}$) while considering the vocabulary size of each NER for unbiased comparison. Note that a large size of vocabulary can have a higher chance of capturing broad keywords. The entity coverage is detailed in Table \ref{tab:ner}, where $N_{g}$ effectively identifies keywords in documents related to the natural conversation and question-answering (QA), while $N_{s}$ adequately finds keywords from professional jargon like medical and scientific claims. 

Based on this analysis, a higher value of $D_{T}$ indicates that a dataset is more similar to a group of general documents, enabling the linking of diverse documents without concerns of domain-specific jargon, resulting in a lower score (i.e., $\delta$). Conversely, a lower $D_{T}$ value suggests that a dataset consists of specialized documents, which benefits from linking similar documents that share domain-specific jargon, resulting in higher scores (i.e., $1-\delta$). Thus, general and specialized documents are considered opposites. In Section \ref{sec:experiment}, we tested the UDL across multiple datasets from different domains (e.g., QA, scientific documents) to show its applicability without requiring a specific NER for each domain. This was confirmed with the selected NERs but our UDL could be readily extendable to any other NER.

Finally, in step \textbf{C}, we calculate the cosine similarity between documents based on the model from step \textbf{A} and establish links when the similarity surpasses a score from step \textbf{B}. 
% In Section \ref{sec:experiment} and Appendix \ref{app:ablation}, we covered the ablation studies of UDL.

\begin{table}
\centering
\caption{Query augmentations with Distilled-BERT. Performances (SD) are from NFCorpus, SciFact, ArguAna.}
\scalebox{0.605}{
% \resizebox{6.0cm}{!}{%{1.1cm}{
\begin{tabular}{c|c|c|c} 
\toprule
\textbf{Method}                    & \textbf{N@10} & \textbf{R@100} & \textbf{\# Parameters}  \\ 
\hline
Off-the-shelf       & 40.7 (0.0)                                                                & 67.5 (0.0)                                                                   & -                       \\ 
\hline
\rowcolor{Gray} Cropping  \cite{izacard2021contriever}           & 38.8 (0.4)                                                              & 68.3 (0.5)                                                                 & -                       \\ 
\hline
\rowcolor{Gray} RM3 \cite{inproceedings}           & 41.7 (0.4)                                                              & 70.2 (0.4)                                                                 & -                       \\ 
\hline
\rowcolor{Gray} AxiomaticQE \cite{10.1007/978-3-030-15712-8_24}           & 43.4 (0.5)                                                              & 69.7 (0.3)                                                                 & -                       \\ 
\hline
\rowcolor{Gray} Summarization \cite{zhang2019pegasus}      & 43.3 (0.6)                                                              & 69.4 (0.2)                                                                 & 569M                    \\ 
\hline
\rowcolor{Gray} Flan \cite{flan}               & 44.3 (0.3)                                                              & 70.4 (0.3)                                                                 & 248M                    \\ 
\hline
\rowcolor{Gray} OpenLLaMA \cite{openlm2023openllama}         & 47.0 (0.4)                                                                & 72.5 (0.5)                                                                 & 3B                      \\ 
\hline
\rowcolor{Gray} QGen \cite{raffel2020exploring}               & 46.3 (0.5)                                                              & 71.9 (0.4)                                                                 & 109M                    \\ 
\hline
\rowcolor{LightCyan} UDL +~RM3            & 44.0 (0.4)                                                              & 71.6 (0.5)                                                                 & 109M                       \\ 
\hline
\rowcolor{LightCyan} UDL +~AxiomaticQE            & 44.5 (0.3)                                                              & 71.4 (0.5)                                                                 & 109M                       \\ 
\hline
\rowcolor{LightCyan} UDL +~Summarization & 45.1 (0.4)                                                              & 71.7 (0.4)                                                                 & 678M                    \\ 
\hline
\rowcolor{LightCyan} UDL +~Flan          & 45.2 (0.6)                                                              & 72.1 (0.5)                                                                 & 357M                    \\ 
\hline
\rowcolor{LightCyan} UDL +~OpenLLaMA     & 48.2 (0.2)                                                              & 73.1 (0.3)                                                                 & 3.1B                    \\ 
\hline
\rowcolor{LightCyan} UDL +~QGen          & \textbf{49.5 (0.3)}                                                              & \textbf{73.6 (0.4)}                                                                 & 218M                    \\
\hline
\rowcolor{celadon} Mapping +~QGen         & 47.6 (0.4)                                                              & 72.6 (0.5)                                                                 & 218M                    \\
% \hline
% \rowcolor{celadon} GloVe +~QGen         & 47.2 (0.4)                                                              & 72 (0.5)                                                                 & 218M                    \\
\hline
\rowcolor{celadon} TF-IDF +~QGen         & 47.7 (0.5)                                                              & 72.9 (0.5)                                                                 & 218M                    \\
\hline
\rowcolor{celadon} LM \cite{song2020mpnet} +~QGen         & 48.2 (0.3)                                                              & 72.7 (0.3)                                                                 & 218M                    \\
\hline
\rowcolor{celadon} Fixed score (0.4) +~QGen         & 46.9 (0.4)                                                              & 72.1 (0.4)                                                                 & 218M                    \\
\hline
\rowcolor{celadon} Fixed score (0.6) +~QGen         & 47.8 (0.2)                                                              & 72.5 (0.4)                                                                 & 218M                    \\

\bottomrule
\end{tabular}
}
\label{tab:query_gen}
\end{table}

\begin{figure*}[h]
\centering
\includegraphics[width=0.73\linewidth]{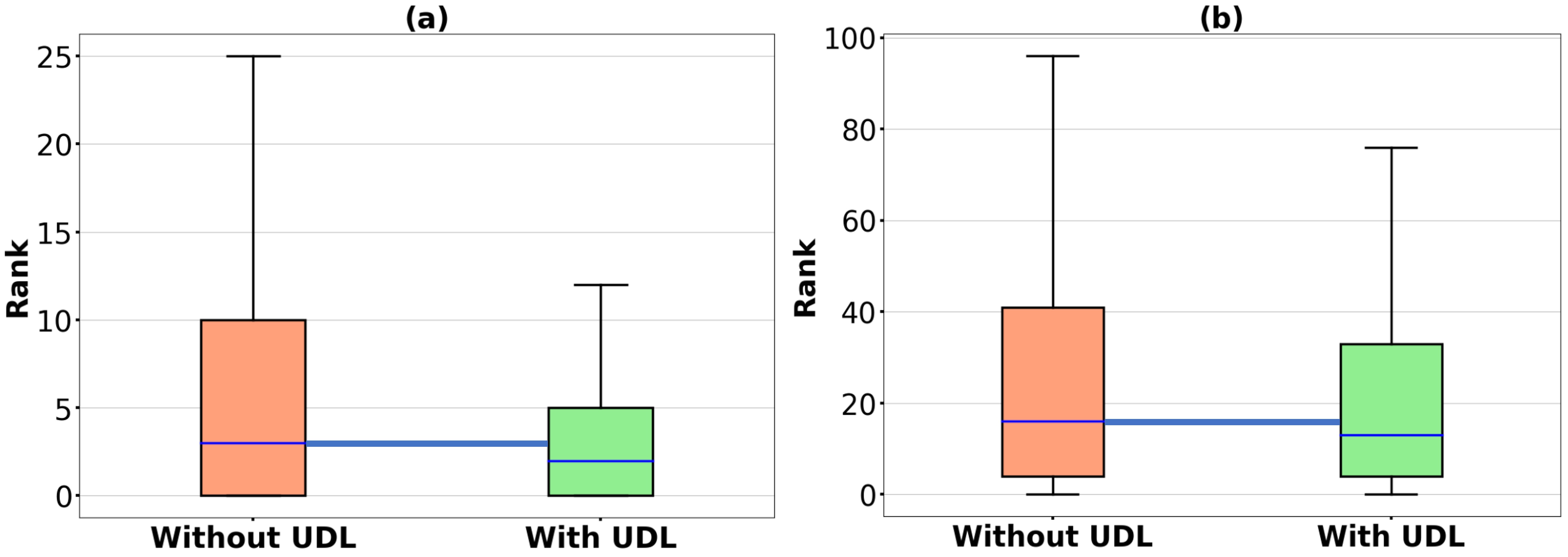} 
\caption{Distribution of rank of correctly classified queries when $k$=100 in NFCorpus, SciFact, ArguAna. (a) Single linked query-document. (b) Multiple linked query-documents. Blue line: Median value.}
\label{fig:rank_dist}
\end{figure*}

% Table \ref{tab:similarity_model_appendix} empirically confirms the usefulness of our decision of similarity model.
% Table \ref{tab:threshold_appendix} shows the practicality of our decision of similarity score.

% We experimented UDL on nine datasets with eight pre-trained IR models and compared with six SOTA models. For the UDL, we linked the two closest documents by concatenation and generated three synthetic queries for each linked and unlinked document. We empirically decided $\gamma$=70 and $\delta$=0.4. trained the IR model three times with different random seeds and reported the averaged NDCG@10 (N@10) and Recall@100 (R@100), along with the standard deviation (SD). 

 % touvron2023llama
\section{Results and Discussions}
\label{sec:experiment}
\noindent
\textbf{Experimental Setup}\;\;
The details of the experimental setup are covered in Appendix \ref{app:setup}, where
% where we cover the representative results and compare with multiple state-of-the-art (SOTA) models in this section.
% Further experiments in non-English and shopping queries are shown in Appendix \ref{sess:add_experiments}. 
we empirically set two hyperparameters in UDL as $\gamma$=0.7 and $\delta$=0.4, and reported the averaged NDCG@$k$ (N@$k$) and Recall@$k$ (R@$k$), along with the standard deviation (SD). For reproducibility, the training framework is covered in Appendix \ref{app:comp}, and the code is included in the supplementary material. Steps of fine-tuning are as follows: \textbf{(1)} Classifying linked and unlinked documents based on UDL, taking into account the order of the linked ones. \textbf{(2)} Feeding them as the inputs to the models and generating the synthetic queries with the same process as the original approach (e.g., model or prompt-based generations). \textbf{(3)} Fine-tuning the IR models based on generated queries.
% Assigning the used documents as relevant to the generated synthetic queries and fine-tuning IR models.
%  where more details can be found in Appendix \ref{app:setup}

\noindent
\textbf{Research Questions}\;\;
We aim to address four research questions (RQs): %based on concrete experiments and studies 
\textbf{RQ1.} What is the most suitable query augmentation method in zero-shot IR?
\textbf{RQ2.} How does UDL enhance zero-shot IR?
\textbf{RQ3.} How well does UDL generalize?
\textbf{RQ4.} Is UDL competitive with state-of-the-art (SOTA)?

\noindent
\textbf{Main Results}\;\;  
Table \ref{tab:query_gen} shows averaged results based on different query augmentations where we generated the same number of queries for each method.
% where we consider the random cropping \citet{izacard2021contriever}, RM3 \cite{inproceedings}, AxiomaticQE \cite{10.1007/978-3-030-15712-8_24}, summarization \citet{zhang2019pegasus}, pseudo query generations with QGen \citet{raffel2020exploring}, Flan \citet{flan} and OpenLLaMA \citet{openlm2023openllama, together2023redpajama}. 
% UDL significantly improves every baseline we've tested.  
%Combining powerful LM techniques with UDL yields better performance (\textbf{RQ1}). 
The overall trend of LM-based approaches outperforming simpler methods persists when UDL is added. However, a relatively parameter-efficient combination of UDL+QGen (218M) showed the best performance overall (\textbf{RQ1}), outperforming UDL+OpenLLaMA (3.1B). This promises significant savings of computational resources at scale. From our initial investigation, we found that OpenLLaMA tends to become more verbose after incorporating UDL, which may increase the risk of hallucination. In contrast, QGen generates more concise queries that are likely more accurate and relevant to the document. Additionally, we did not modify the LLM prompts based on UDL in this work, which presents a valuable future direction to optimize the prompts to better cover linked ones.

%when applied at scale.

%Also, implementing the UDL demonstrates further improvement, highlighting the extensibility of our approach (\textbf{RQ1}). UDL is efficient: UDL+QGen outperforms OpenLLaMA with fewer parameters (218M vs 3B), saving computational resources.
%Lastly, we tested the effectiveness of synthetic queries generated from linked documents. For this, we generated the synthetic query in a single document and mapped the multiple documents decided as the linking from UDL (i.e., Mapping+QGen). This enhances the performance of QGen, meaning that our UDL properly links similar documents. Further improvement can be gained from UDL which confirms the usefulness of generating synthetic queries from linked documents.  

Furthermore, we ablated the document merging mechanism of UDL by generating the synthetic queries from each document individually and mapping them to documents found by the linking procedure (Mapping+QGen in Table \ref{tab:query_gen}). While this still outperformed the corresponding baseline (QGen), it performed worse than complete UDL. This suggests that generating queries from the merged documents improves model generalization by introducing harder queries with increased ambiguity compared to the original. Indeed, Table~\ref{tab:example_query} anecdotally shows that resulting queries fit both linked documents and are generally less specific. Besides, the linking mechanism itself provides a more exhaustive way of identifying positive query-document pairs, improving the performance (\textbf{RQ2}). Figure~\ref{fig:rank_dist} illustrates this behavior: Distributions with UDL are more compact, have fewer outliers, and allocate higher ranks for relevant documents.

Lastly, we investigated the influence of decisions in UDL separately. We compared the results between fixed similarity models (i.e., TF-IDF or LM+QGen) and flexible ones (i.e., UDL+QGen) where the latter excels. Also, we tested the results by fixing the similarity scores (i.e., Fixed score (0.4) or Fixed score (0.6)+QGen) and LM where flexible scores from UDL enhances the performance. Therefore, our evolved approach with flexible choices of the similarity models and scores promises the results.
% More details are covered in Appendix \ref{app:comp}.

% where the former has 218M parameters and required about 1682 seconds for 10K documents with single NVIDIA A100 while the latter consists of B parameters and took about 12542 seconds in same environment. 

% OpenLLaMA which has enormous parameters (3B) shows the best result while LLM particularly for query generation (i.e. QGen) gives a similar result with smaller parameters (390M). Thus, QGen can be the best choice for zero-shot IR to save both resources and performances  . Finally, using QGen in conjunction with the UDL demonstrates further improvement which highlights the usefulness of our approach.

% (i.e. summarization, QGen and Flan) 

\begin{figure}[!b]
\centering
\includegraphics[width=0.7\linewidth]{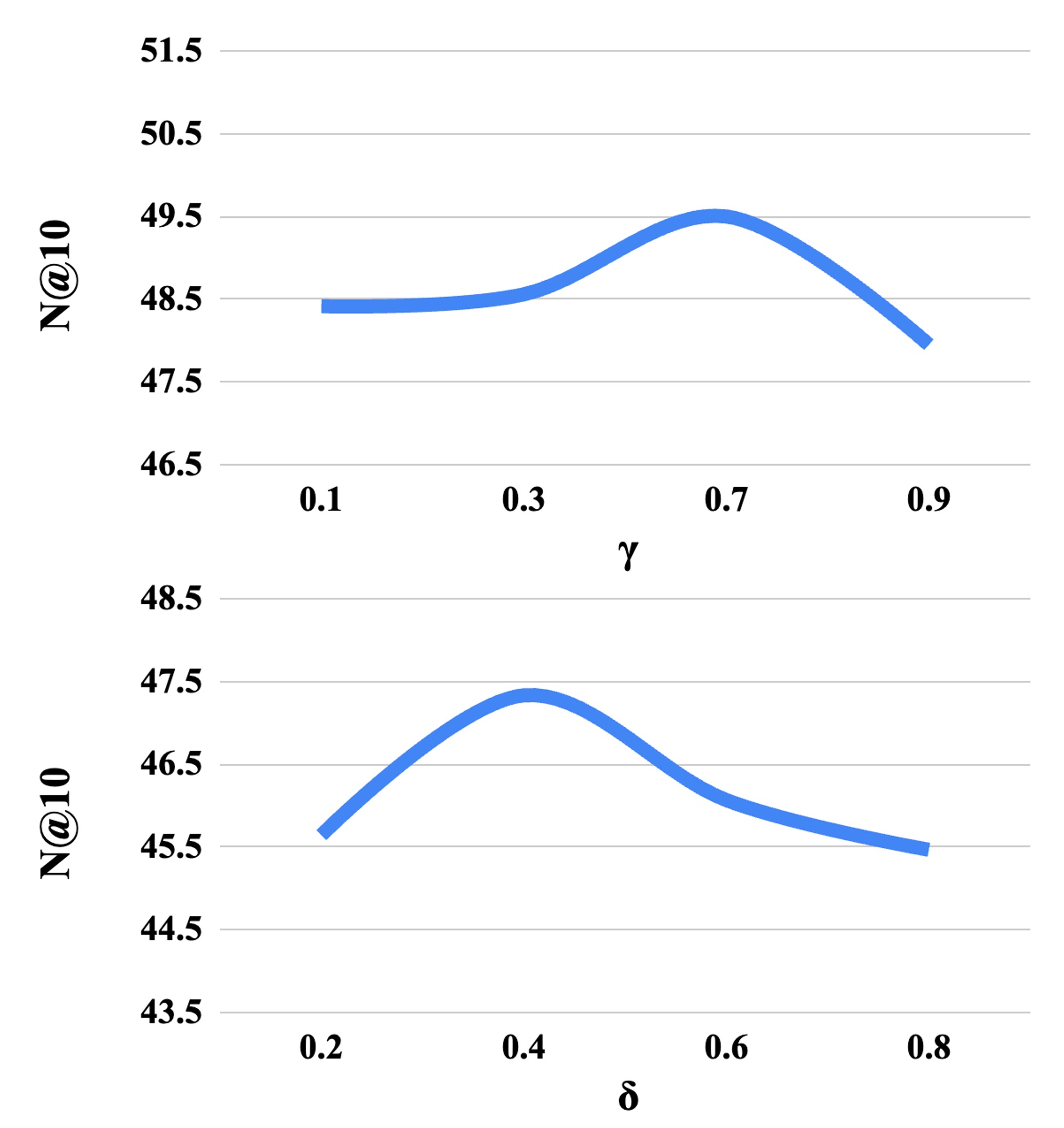} 
\caption{Grid search for $\gamma$ and $\delta$.}
\label{fig:gamma_delta}
\end{figure}

% \vspace{-1.3cm}
\noindent
\textbf{Hyperparameters Choice}\;\; 
%To assess the effectiveness of UDL, we examined examples of synthetic queries generated by UDL in Table \ref{tab:example_query}. It's evident that synthetic queries generated with UDL represent the intersection between two documents. 
%==OLD SENTENCE==; rewrote because of the complaint of reviewer GNif:
% This not only aids in locating relevant documents across multiple linked query-documents by incorporating the generated similar cases during fine-tuning, but also enhances search within a single linked query-document by presenting broad and ambiguous synthetic queries, thereby improving generalization. 
%This improves model generalization by introducing harder queries with increased ambiguity compared to the original. On the other hand, this increase is controlled since each of the linked documents will be appropriately labeled as positive for the corresponding synthetic query. 
%This observation is supported by Figure \ref{fig:rank_dist}, where UDL assigns higher ranks to relevant documents. Thus, the primary impact of UDL lies in generating synthetic queries that span multiple documents, facilitating the retrieval of relevant documents for the test query with higher priority (\textbf{RQ2}). 
%  and can also be confirmed from the comparison between Mapping+QGen and UDL+QGen in Table \ref{tab:query_gen}
Figure \ref{fig:gamma_delta} shows the grid search for UDL's hyperparameters using NFCorpus, SciFact, and ArguAna yielding $\gamma$=0.7 and $\delta$=0.4 as most optimal. (see Tables~\ref{tab:similarity_model_appendix} and \ref{tab:threshold_appendix} for detailed results). We also checked the quality between synthetic queries and the offered train queries in used datasets. Detail of logic is shown in Algorithm \ref{algo:quality} where 93\% of synthetic queries generated from linked documents in UDL have sufficient quality as the train set to map the relevant documents. 
 
% shows the distribution of rank of classified test queries where 
% (i.e. italic font below each dataset)
% [MAYBE WE CAN MOVE TO APPENDIX]
% In Table \ref{tab:ablation_udl}, we investigated two datasets with different similarity models and scores as an ablation study. We can find that the decisions from UDL promise better performances. More investigations are shown in Tables \ref{tab:similarity_model_appendix} and \ref{tab:threshold_appendix} where similar patterns are observed in other datasets showing the empirical improvements using the UDL. 

\begin{table*}[!t]
\centering
\caption{Performances in English datasets. $\dagger$: In-domain result since Quora was exposed for pre-training before fine-tuning with UDL. SD is always lower than 0.7. QGen and UDL+QGen have same number of generated queries.}
\resizebox{14.5cm}{!}{
\begin{tabular}{c|c|c|c|c|c|c|c|c|c|c|c|c|c|c|c} 
\toprule
                                 & \textbf{Data}                            & \multicolumn{2}{c|}{NFCorpus}                                                         & \multicolumn{2}{c|}{SciFact}                                                          & \multicolumn{2}{c|}{ArguAna}                                                          & \multicolumn{2}{c|}{SCIDOCS}                                                          & \multicolumn{2}{c|}{Climate-FEVER}                                                    & \multicolumn{2}{c|}{TREC-COVID}                                                       & \multicolumn{2}{c}{Quora}                                                                                \\ 
\hline
\textbf{Model}                   &                                          & N@10                                      & R@100                                     & N@10                                      & R@100                                     & N@10                                      & R@100                                     & N@10                                      & R@100                                     & N@10                                      & R@100                                     & N@10                                      & R@100                                     & N@10                                               & R@100                                               \\ 
\hline
\multirow{3}{*}{All-MPNet}       & Off-the-shelf                            & 33.3                                      & 33.9                                      & 65.6                                      & 94.2                                      & 46.5                                      & 98.7                                      & \textbf{23.8}                             & \textbf{55.0}                             & 22.0                                      & 54.5                                      & 51.3                                      & 10.6                                      & 87.5$\dagger$                                      & 99.6$\dagger$                                       \\
                                 & {\cellcolor[rgb]{0.9,0.9,0.9}}QGen & {\cellcolor[rgb]{0.9,0.9,0.9}}33.1  & {\cellcolor[rgb]{0.9,0.9,0.9}}31.3  & {\cellcolor[rgb]{0.9,0.9,0.9}}65.2  & {\cellcolor[rgb]{0.9,0.9,0.9}}91.6  & {\cellcolor[rgb]{0.9,0.9,0.9}}53.3  & {\cellcolor[rgb]{0.9,0.9,0.9}}98.8  & {\cellcolor[rgb]{0.9,0.9,0.9}}19.1  & {\cellcolor[rgb]{0.9,0.9,0.9}}44.4  & {\cellcolor[rgb]{0.9,0.9,0.9}}23.8  & {\cellcolor[rgb]{0.9,0.9,0.9}}54.9  & {\cellcolor[rgb]{0.9,0.9,0.9}}59.8  & {\cellcolor[rgb]{0.9,0.9,0.9}}10.8  & {\cellcolor[rgb]{0.9,0.9,0.9}}86.0$\dagger$  & {\cellcolor[rgb]{0.9,0.9,0.9}}99.2$\dagger$   \\
                                 & {\cellcolor[rgb]{0.878,1,1}}UDL + QGen   & {\cellcolor[rgb]{0.878,1,1}}\textbf{35.9} & {\cellcolor[rgb]{0.878,1,1}}\textbf{34.9} & {\cellcolor[rgb]{0.878,1,1}}\textbf{67.1} & {\cellcolor[rgb]{0.878,1,1}}\textbf{94.8} & {\cellcolor[rgb]{0.878,1,1}}\textbf{61.0} & {\cellcolor[rgb]{0.878,1,1}}\textbf{99.5} & {\cellcolor[rgb]{0.878,1,1}}22.5          & {\cellcolor[rgb]{0.878,1,1}}51.3          & {\cellcolor[rgb]{0.878,1,1}}\textbf{24.1} & {\cellcolor[rgb]{0.878,1,1}}\textbf{55.4} & {\cellcolor[rgb]{0.878,1,1}}\textbf{69.5} & {\cellcolor[rgb]{0.878,1,1}}\textbf{12.2} & {\cellcolor[rgb]{0.878,1,1}}\textbf{88.1}$\dagger$ & {\cellcolor[rgb]{0.878,1,1}}\textbf{99.7}$\dagger$  \\ 
\hline
\multirow{3}{*}{Distilled-BERT}  & Off-the-shelf                            & 25.6                                      & 23.3                                      & 53.8                                      & 84.6                                      & 42.6                                      & 94.6                                      & 13.3                                      & 29.7                                      & 20.2                                      & 44.6                                      & 47.8                                      & 7.2                                       & 85.5                                               & 98.9                                                \\
                                 & {\cellcolor[rgb]{0.9,0.9,0.9}}QGen & {\cellcolor[rgb]{0.9,0.9,0.9}}29.0  & {\cellcolor[rgb]{0.9,0.9,0.9}}27.1  & {\cellcolor[rgb]{0.9,0.9,0.9}}59.6  & {\cellcolor[rgb]{0.9,0.9,0.9}}90.1  & {\cellcolor[rgb]{0.9,0.9,0.9}}50.3  & {\cellcolor[rgb]{0.9,0.9,0.9}}98.5  & {\cellcolor[rgb]{0.9,0.9,0.9}}14.4  & {\cellcolor[rgb]{0.9,0.9,0.9}}33.1  & {\cellcolor[rgb]{0.9,0.9,0.9}}22.0  & {\cellcolor[rgb]{0.9,0.9,0.9}}52.3  & {\cellcolor[rgb]{0.9,0.9,0.9}}56.9  & {\cellcolor[rgb]{0.9,0.9,0.9}}9.8   & {\cellcolor[rgb]{0.9,0.9,0.9}}84.5           & {\cellcolor[rgb]{0.9,0.9,0.9}}98.7            \\
                                 & {\cellcolor[rgb]{0.878,1,1}}UDL + QGen   & {\cellcolor[rgb]{0.878,1,1}}\textbf{31.2} & {\cellcolor[rgb]{0.878,1,1}}\textbf{30.8} & {\cellcolor[rgb]{0.878,1,1}}\textbf{61.5} & {\cellcolor[rgb]{0.878,1,1}}\textbf{90.7} & {\cellcolor[rgb]{0.878,1,1}}\textbf{55.8} & {\cellcolor[rgb]{0.878,1,1}}\textbf{99.2} & {\cellcolor[rgb]{0.878,1,1}}\textbf{16.6} & {\cellcolor[rgb]{0.878,1,1}}\textbf{40.5} & {\cellcolor[rgb]{0.878,1,1}}\textbf{22.3} & {\cellcolor[rgb]{0.878,1,1}}\textbf{52.8} & {\cellcolor[rgb]{0.878,1,1}}\textbf{61.7} & {\cellcolor[rgb]{0.878,1,1}}\textbf{10.9} & {\cellcolor[rgb]{0.878,1,1}}\textbf{85.8}          & {\cellcolor[rgb]{0.878,1,1}}\textbf{99.1}           \\ 
\hline
\multirow{3}{*}{SGPT}            & Off-the-shelf                            & 21.7                                      & 23.3                                      & 54.3                                      & 85.7                                      & 41.1                                      & 94.6                                      & 11.7                                      & 26.9                                      & 20.8                                      & 45.5                                      & 57.2                                      & 9.3                                       & 81.7                                               & 97.8                                                \\
                                 & {\cellcolor[rgb]{0.9,0.9,0.9}}QGen & {\cellcolor[rgb]{0.9,0.9,0.9}}24.1  & {\cellcolor[rgb]{0.9,0.9,0.9}}23.8  & {\cellcolor[rgb]{0.9,0.9,0.9}}56.8  & {\cellcolor[rgb]{0.9,0.9,0.9}}88.9  & {\cellcolor[rgb]{0.9,0.9,0.9}}47.4  & {\cellcolor[rgb]{0.9,0.9,0.9}}96.9  & {\cellcolor[rgb]{0.9,0.9,0.9}}12.6  & {\cellcolor[rgb]{0.9,0.9,0.9}}29.8  & {\cellcolor[rgb]{0.9,0.9,0.9}}21.1  & {\cellcolor[rgb]{0.9,0.9,0.9}}48.0  & {\cellcolor[rgb]{0.9,0.9,0.9}}61.6  & {\cellcolor[rgb]{0.9,0.9,0.9}}9.5   & {\cellcolor[rgb]{0.9,0.9,0.9}}83.9           & {\cellcolor[rgb]{0.9,0.9,0.9}}98.6            \\
                                 & {\cellcolor[rgb]{0.878,1,1}}UDL + QGen   & {\cellcolor[rgb]{0.878,1,1}}\textbf{24.6} & {\cellcolor[rgb]{0.878,1,1}}\textbf{26.0} & {\cellcolor[rgb]{0.878,1,1}}\textbf{57.4} & {\cellcolor[rgb]{0.878,1,1}}\textbf{90.0} & {\cellcolor[rgb]{0.878,1,1}}\textbf{52.0} & {\cellcolor[rgb]{0.878,1,1}}\textbf{99.1} & {\cellcolor[rgb]{0.878,1,1}}\textbf{15.3} & {\cellcolor[rgb]{0.878,1,1}}\textbf{37.1} & {\cellcolor[rgb]{0.878,1,1}}\textbf{21.5} & {\cellcolor[rgb]{0.878,1,1}}\textbf{48.4} & {\cellcolor[rgb]{0.878,1,1}}\textbf{64.5} & {\cellcolor[rgb]{0.878,1,1}}\textbf{10.6} & {\cellcolor[rgb]{0.878,1,1}}\textbf{85.0}          & {\cellcolor[rgb]{0.878,1,1}}\textbf{99.0}           \\ 
\hline
\multirow{3}{*}{M-Distilled USE} & Off-the-shelf                            & 20.0                                      & 24.2                                      & 39.0                                      & 74.7                                      & 48.7                                      & 97.1                                      & 9.3                                       & 27.5                                      & 13.0                                      & 37.5                                      & 23.9                                      & 3.5                                       & 82.4                                               & 98.4                                                \\
                                 & {\cellcolor[rgb]{0.9,0.9,0.9}}QGen & {\cellcolor[rgb]{0.9,0.9,0.9}}24.8  & {\cellcolor[rgb]{0.9,0.9,0.9}}24.7  & {\cellcolor[rgb]{0.9,0.9,0.9}}48.9  & {\cellcolor[rgb]{0.9,0.9,0.9}}81.9  & {\cellcolor[rgb]{0.9,0.9,0.9}}47.9  & {\cellcolor[rgb]{0.9,0.9,0.9}}97.3  & {\cellcolor[rgb]{0.9,0.9,0.9}}13.5  & {\cellcolor[rgb]{0.9,0.9,0.9}}32.0  & {\cellcolor[rgb]{0.9,0.9,0.9}}16.3  & {\cellcolor[rgb]{0.9,0.9,0.9}}40.0  & {\cellcolor[rgb]{0.9,0.9,0.9}}57.0  & {\cellcolor[rgb]{0.9,0.9,0.9}}10.6  & {\cellcolor[rgb]{0.9,0.9,0.9}}83.4           & {\cellcolor[rgb]{0.9,0.9,0.9}}98.6            \\
                                 & {\cellcolor[rgb]{0.878,1,1}}UDL + QGen   & {\cellcolor[rgb]{0.878,1,1}}\textbf{26.9} & {\cellcolor[rgb]{0.878,1,1}}\textbf{27.9} & {\cellcolor[rgb]{0.878,1,1}}\textbf{49.9} & {\cellcolor[rgb]{0.878,1,1}}\textbf{84.1} & {\cellcolor[rgb]{0.878,1,1}}\textbf{49.1} & {\cellcolor[rgb]{0.878,1,1}}\textbf{98.5} & {\cellcolor[rgb]{0.878,1,1}}\textbf{15.1} & {\cellcolor[rgb]{0.878,1,1}}\textbf{38.3} & {\cellcolor[rgb]{0.878,1,1}}\textbf{16.7} & {\cellcolor[rgb]{0.878,1,1}}\textbf{42.7} & {\cellcolor[rgb]{0.878,1,1}}\textbf{62.0} & {\cellcolor[rgb]{0.878,1,1}}\textbf{11.5} & {\cellcolor[rgb]{0.878,1,1}}\textbf{84.3}          & {\cellcolor[rgb]{0.878,1,1}}\textbf{99.0}           \\
\bottomrule
\end{tabular}}
\label{tab:result}
\end{table*}

\begin{table}
\centering
\caption{Performances in non-English datasets where SD is always lower than 0.7.}
\resizebox{7cm}{!}{
\begin{tabular}{c|c|c|c|c|c} 
\toprule
                                          & \textbf{\textbf{Data}}                   & \multicolumn{2}{c|}{ViHealthQA}                                                                         & \multicolumn{2}{c}{GermanQuAD}                                                                           \\ 
\hline
\textbf{\textbf{\textbf{\textbf{Model}}}} &                                          & N@10                                               & R@100                                              & N@10                                               & R@100                                               \\ 
\hline
\multirow{3}{*}{M-Distilled~USE}          & Off-the-shelf                            & 9.3                                                & 21.6                                               & 33.4                                               & 67.0                                                \\
                                          & {\cellcolor[rgb]{0.9,0.9,0.9}}QGen & {\cellcolor[rgb]{0.9,0.9,0.9}}22.2           & {\cellcolor[rgb]{0.9,0.9,0.9}}33.8           & {\cellcolor[rgb]{0.9,0.9,0.9}}31.7           & {\cellcolor[rgb]{0.9,0.9,0.9}}65.8            \\
                                          & {\cellcolor[rgb]{0.878,1,1}}UDL + QGen   & {\cellcolor[rgb]{0.878,1,1}}\textbf{23.0}          & {\cellcolor[rgb]{0.878,1,1}}\textbf{\textbf{34.8}} & {\cellcolor[rgb]{0.878,1,1}}\textbf{34.7}          & {\cellcolor[rgb]{0.878,1,1}}\textbf{69.0}           \\ 
\hline
\multirow{3}{*}{V-SBERT}                  & Off-the-shelf                            & 13.8                                               & 27.6                                               & -                                                  & -                                                   \\
                                          & {\cellcolor[rgb]{0.9,0.9,0.9}}QGen & {\cellcolor[rgb]{0.9,0.9,0.9}}22.9           & {\cellcolor[rgb]{0.9,0.9,0.9}}33.6           & {\cellcolor[rgb]{0.9,0.9,0.9}}-              & {\cellcolor[rgb]{0.9,0.9,0.9}}-               \\
                                          & {\cellcolor[rgb]{0.878,1,1}}UDL + QGen   & {\cellcolor[rgb]{0.878,1,1}}\textbf{\textbf{23.8}} & {\cellcolor[rgb]{0.878,1,1}}\textbf{\textbf{34.8}} & {\cellcolor[rgb]{0.878,1,1}}-                      & {\cellcolor[rgb]{0.878,1,1}}-                       \\ 
\hline
\multirow{3}{*}{V-SimeCSE~}               & Off-the-shelf                            & 10.9                                               & 23.4                                               & -                                                  & -                                                   \\
                                          & {\cellcolor[rgb]{0.9,0.9,0.9}}QGen & {\cellcolor[rgb]{0.9,0.9,0.9}}22.5           & {\cellcolor[rgb]{0.9,0.9,0.9}}33.4           & {\cellcolor[rgb]{0.9,0.9,0.9}}-              & {\cellcolor[rgb]{0.9,0.9,0.9}}-               \\
                                          & {\cellcolor[rgb]{0.878,1,1}}UDL + QGen   & {\cellcolor[rgb]{0.878,1,1}}\textbf{23.4}          & {\cellcolor[rgb]{0.878,1,1}}\textbf{34.6}          & {\cellcolor[rgb]{0.878,1,1}}-                      & {\cellcolor[rgb]{0.878,1,1}}-                       \\ 
\hline
\multirow{3}{*}{G-Electra~}               & Off-the-shelf                            & -                                                  & -                                                  & 25.0                                               & 53.5                                                \\
                                          & {\cellcolor[rgb]{0.9,0.9,0.9}}QGen & {\cellcolor[rgb]{0.9,0.9,0.9}}-              & {\cellcolor[rgb]{0.9,0.9,0.9}}-              & {\cellcolor[rgb]{0.9,0.9,0.9}}28.1           & {\cellcolor[rgb]{0.9,0.9,0.9}}59.7            \\
                                          & {\cellcolor[rgb]{0.878,1,1}}UDL + QGen   & {\cellcolor[rgb]{0.878,1,1}}-                      & {\cellcolor[rgb]{0.878,1,1}}-                      & {\cellcolor[rgb]{0.878,1,1}}\textbf{30.6}          & {\cellcolor[rgb]{0.878,1,1}}\textbf{60.8}           \\ 
\hline
\multirow{3}{*}{G-XLM-R}                  & Off-the-shelf                            & -                                                  & -                                                  & 8.3                                                & 24.7                                                \\
                                          & {\cellcolor[rgb]{0.9,0.9,0.9}}QGen & {\cellcolor[rgb]{0.9,0.9,0.9}}-              & {\cellcolor[rgb]{0.9,0.9,0.9}}-              & {\cellcolor[rgb]{0.9,0.9,0.9}}36.0           & {\cellcolor[rgb]{0.9,0.9,0.9}}70.5            \\
                                          & {\cellcolor[rgb]{0.878,1,1}}UDL + QGen   & {\cellcolor[rgb]{0.878,1,1}}-                      & {\cellcolor[rgb]{0.878,1,1}}-                      & {\cellcolor[rgb]{0.878,1,1}}\textbf{\textbf{36.6}} & {\cellcolor[rgb]{0.878,1,1}}\textbf{\textbf{71.2}}  \\
\bottomrule
\end{tabular}}
\label{tab:result_multi}
\end{table}

\noindent
\textbf{Does UDL generalize?}\;\; 
Table \ref{tab:result} compares the results of off-the-shelf models to those that have been fine-tuned across various models and English datasets. Interestingly, fine-tuning with QGen does not always improve the results, especially in high-performance models (e.g., All-MPNet). This suggests that synthetic queries can potentially decrease domain adaptation. Generally, we observe further improvements with UDL, except for SCIDOCS with All-MPNet. In such cases, UDL remains superior to naive fine-tuning. 
Table \ref{tab:result_multi} demonstrates the results of UDL compared to the off-the-shelf models in Vietnamese and German datasets. The findings show the superiority of UDL when applied to non-English languages which confirms the flexibility of UDL. Table \ref{tab:ma_amazon} covers the results in MA-Amazon \cite{reddy2022shopping} with our approach and compares them with SOTA. This dataset contains interactions between user search queries and product information, along with relevance labels, making it well-suited for evaluating the extensibility of our method in real-world scenarios. Similar to the previous experiments, QGen improves the zero-shot performances where it is further enhanced consistently with the UDL approach. Therefore, our UDL is still generalized properly in potential real-world implementations. Even if SOTA models have bigger sizes and access to real user queries for pre-training and fine-tuning, the combination of UDL and QGen outperforms them significantly. Note that SOTA models consist of larger parameters and utilize the 482K unique documents for pre-training and 17K query-document pairs for fine-tuning. This confirms both the cost-effectiveness and resource-effectiveness of the UDL to achieve better performance than SOTA. Thus, we can verify that UDL works well across multiple datasets, languages, and models (\textbf{RQ3}).

% \noindent
% \textbf{Shopping Queries Dataset}\;\;
% Table \ref{tab:ma_amazon} shows the results in MA-Amazon dataset with our approach and compares them with SOTA. Similar to the previous experiments, QGen improves the zero-shot performances where it is further enhanced consistently with the UDL approach. Therefore, our UDL is still generalized properly in potential real-world implementation. Even if SOTA models have bigger sizes and access to real user queries for pre-training and fine-tuning, the combination of UDL and QGen outperforms them significantly. Note that SOTA models consist of larger parameters and utilize the 482K unique documents for pre-training and 17K query-document pairs for fine-tuning. This confirms both the cost-effectiveness and resource-effectiveness of the UDL to achieve better performance than SOTA. 

\begin{table}
\centering
\caption{Performances in shopping query dataset where SD in Distilled-BERT is always under 0.4. SOTA results are exported from \citet{sun2023pretraining}.}  
\resizebox{7.6cm}{!}{
\begin{tabular}{c|c|c|c|c|c|c} 
\toprule
\textbf{\textbf{\textbf{\textbf{Model}}}} & \textbf{Method}                                                                      & \textbf{Data}                                                              & \textbf{N@50} & \textbf{R@100} & \textbf{R@500} & \multicolumn{1}{l}{\textbf{\# Parameters}}  \\ 
\hline
\multirow{3}{*}{Distilled-BERT}           & Off-the-shelf                                                                        & \multirow{3}{*}{Document}                                                  & 39.0            & 57.8           & 73.5           & \multirow{3}{*}{66M}                        \\
                                          & \cellcolor{Gray} QGen                                                                                 &                                                                            & \cellcolor{Gray}43.5          & \cellcolor{Gray}65.2           & \cellcolor{Gray}80.6           &                                             \\
                                          & \cellcolor{LightCyan} UDL + QGen                                                                           &                                                                            & \cellcolor{LightCyan}\textbf{44.6} & \cellcolor{LightCyan}\textbf{66.8}  & \cellcolor{LightCyan}\textbf{82.5}  &                                             \\ 
\hline
BIBERT                                    & \multirow{4}{*}{\begin{tabular}[c]{@{}c@{}}Pre-training +\\Fine-tuning\end{tabular}} & \multirow{4}{*}{\begin{tabular}[c]{@{}c@{}}Query +\\Document\end{tabular}} & 40.1          & 61.4           & 78.1           & \multirow{4}{*}{\textasciitilde{}109M}      \\ 
\cline{1-1}\cline{4-6}
MTBERT                                    &                                                                                      &                                                                            & 40.0            & 61.4           & 78.4           &                                             \\ 
\cline{1-1}\cline{4-6}
MADRAL                                    &                                                                                      &                                                                            & 40.4          & 61.7           & 78.5           &                                             \\ 
\cline{1-1}\cline{4-6}
ATTEMPT                                   &                                                                                      &                                                                            & 41.0            & 62.3           & 79.2           &                                             \\
\bottomrule
\end{tabular}}
\label{tab:ma_amazon}
\end{table}

% Further experiments are in Appendix \ref{sess:add_experiments}.
%, particularly for the ArguAna dataset, which is categorized as a general document based on decision of similarity threshold in UDL. This seems to be because the UDL identifies similar documents more accurately, given that the utilized LM is intensively pre-trained on similar datasets (e.g. Wikipedia)

A comparison between SOTA and QGen with UDL in English datasets is shown in Table \ref{tab:sota_compare} where all IR models have approximately 100M parameters for each encoder. Notably, All-MPNet with UDL wins others, demonstrating the superiority of UDL (\textbf{RQ4}). In the case of UDL implementation, some of the SOTA models were exposed to the documents of the target dataset during pre-training, but our method achieved better results. Lastly, we focused on directly fine-tuning with UDL, which could be extended to other applications like document expansion. This highlights the versatility of UDL for various tasks and models.

\begin{table}
\centering
\caption{Comparison with SOTA in zero-shot scenarios. UDL: Fine-tuning All-MPNet with UDL.} %Averaged results are from used English datasets, except for Quora which is in-domain. %uniCOIL \citet{lin2021brief}, 
% \scalebox{0.68}{
\resizebox{7.7cm}{!}{
% \resizebox{8.45cm}{!}{
\begin{tabular}{c|c|c|c|c|c|c|c|a} 
\toprule
\textbf{Model} & BM25 & TAS-B & \begin{tabular}[c]{@{}c@{}}Contr-\\iever\end{tabular} & \begin{tabular}[c]{@{}c@{}}SPLA-\\DE++\end{tabular} & ANCE & \begin{tabular}[c]{@{}c@{}}COCO-\\DR\end{tabular} & \begin{tabular}[c]{@{}c@{}}DRA-\\GON+\end{tabular} & \textbf{\textbf{UDL}}   \\ 
\hline
N@10     & 40.5 & 38.2  & 40.8                                                  & 44.8                                                & 35.6 & 45.3                                              & 43.8                                               & \textbf{\textbf{46.7}}  \\ 
\hline
R@100    & 50.1 & 51.6  & 54.5                                                  & 53.7                                                & 46.7 & 53.9                                              & 53.4                                               & \textbf{\textbf{58.0}}    \\
\bottomrule
\end{tabular}
}
\label{tab:sota_compare}
\end{table}

\section{Conclusions}
We propose a novel UDL to mitigate the limitations of conventional fine-tuning of IR models in zero-shot. UDL uses entropy and NER to tailor a linking method for each dataset with diverse tasks. Our comprehensive experiments show the effectiveness of UDL across various datasets and models.
% We propose UDL as a new way to address the limitations of conventional fine-tuning of IR models in zero-shot scenarios. This framework consists of the similarity model based on term entropy and similarity score based on NER, which assigns a customized linking method to each dataset with varying tasks. Our extensive experiments show the superiority of UDL across various datasets, languages and models.

% We believe this work is crucial for deploying LLM, especially in the context of Small Language Models.
% Lastly, it can be extended to the recent SOTA retrieval models and document expansion for further improvement. 

% In the future, we will modify the algorithm to reduce the number of hyperparameters involved.

\section{Limitations}
The proposed UDL offers significant advantages as an application. However, there are three possible limitations to consider. Firstly, while we consistently surpassed naive fine-tuning, there is an inherent limit to the enhancements. The performance of the retrieval model is influenced by the quality of synthetic queries. In general, the advanced pseudo-query generation methods manage multiple documents more effectively, indicating a valuable future direction to combine UDL with competitive pseudo-query generation approaches for further improvement. It also highlights the importance of selecting appropriate query augmentation strategies early in the project. Secondly, there is potential to introduce dynamic criteria, such as $\gamma$ and $\delta$ in UDL, which were empirically defined in this study. Adjustments could be made for each candidate document, tailored to the similarities between documents and their types. Lastly, our comprehensive evaluation of UDL spanned ten datasets with diverse domains and languages (see Tables \ref{tab:result} - \ref{tab:ma_amazon}). There is a scope to extend this to larger documents and other languages, which was challenging due to computational resource constraints. These identified limitations present valuable research directions for those considering the proposed UDL in their applications.

\bibliography{main}

\input{appendices}
\end{document}

%% file: appendices.tex
\clearpage
\appendix
\label{sec:appendix}

% MODEL: \cite{dai2022promptagator}
\section{Setup}
\label{app:setup}
\textbf{Databases}\;\; We tested ten datasets where the summary of the database is shown in Table \ref{tab:dataset}: NFCorpus \cite{boteva2016} has automatically extracted relevance judgments for medical documents. SciFact \cite{wadden-etal-2020-fact} consists of expert-annotated scientific claims with abstracts and rationales. ArguAna \cite{wachsmuth-etal-2018-retrieval} contains the pairs of argument and counterargument from the online debate. SCIDOCS \cite{cohan-etal-2020-specter} has seven document-level tasks from citation prediction, document classification, and recommendation. Climate-FEVER \cite{diggelmann2021climatefever} consists of real-world claims regarding climate-change with manually annotated evidence sentences from Wikipedia. TREC-COVID \cite{voorhees2020treccovid} contains the COVID-related topics with a collection of literature articles where biomedical experts measure the relevancy between articles and topics. Quora \cite{quora} is built for identifying the duplicate question which is necessary for a scalable online knowledge-sharing platform. GermanQuAD \cite{moller2021germanquad} is high-quality and human-labeled German dataset which includes the self-sufficient questions with all relevant information. ViHealthQA \cite{nguyen2022spbertqa} consists of health-interested QA in Vietnamese. Multi-Aspect Amazon ESCI Dataset (MA-Amazon) \cite{reddy2022shopping} has user queries for product search and long lists of product information like title, description, brand, color with four relevance labels.

% : Exact (E), Substitute (S), Complement (C), and
% Irrelevant (I). As \cite{sun2023pretraining}, we only considered the English parts and treated Exact as relevant and the other labels as irrelevant for Recall measurement while assigned the gains of E, S, C, I to 1.0, 0.1, 0.01, 0.0, respectively for NDCG measurement. 

\noindent
\textbf{Models}\;\; In this work, we considered the diverse sets of models where the summary of them is covered in Table \ref{tab:model}: For query augmentation, we tested five pre-trained models: PEGASUS (Summarization) \cite{zhang2019pegasus}, T5-Base (QGen) \cite{raffel2020exploring} for English datasets, mT5-Base (QGen) \cite{xue2021mt5} for Vietnamese and German databases, Flan T5-Base (Flan) \cite{flan}, OpenLLaMA \cite{openlm2023openllama, together2023redpajama, touvron2023llama}. 

For retrieval task, eight pre-trained retrieval models are experimented: M-Distilled USE \cite{yang2019multilingual}, All-MPNet \cite{song2020mpnet}, Distilled-BERT \cite{Sanh2019DistilBERTAD}, SGPT \cite{muennighoff2022sgpt}, V-SBERT \cite{phobert}, V-SimeCSE \cite{gao2021simcse}, G-Electra \cite{clark2020electra}, G-XLM-R \cite{conneau2020unsupervised}.

For pre-trained LM in similarity model, we employed three pre-trained models: All-MPNet \cite{song2020mpnet} for English datasets, V-SBERT \cite{phobert} for Vietnamese database, G-BERT \cite{chan-etal-2020-germans} for German dataset.

For comparison, ten SOTA models are investigated: TAS-B \cite{hofstätter2021efficiently}, Contriever \cite{izacard2021contriever}, SPLADE++ \cite{formal2022distillation}, ANCE \cite{xiong2020approximate}, COCO-DR \cite{yu2022cocodr}, DRAGON+ \cite{lin2023train}, BIBERT \cite{lin2022pretrained}, MTBERT \cite{kong2022multi}, MADRAL \cite{kong2022multi}, ATTEMPT \cite{sun2023pretraining}.

Table \ref{tab:ner} describes the details of NER models used in this work. NER model trained with general sources ($N_{g}$) covers the diverse types of general entities while NER model trained with specialized sources ($N_{s}$) addresses the various types of medical and scientific entities mostly related to the jargon.

\begin{table}
\centering
\caption{Details of datasets used where we only cover the size of test set which is our point of interest. Note that ViHealthQA did not report the licenses in the paper or a repository.}
\scalebox{0.6}{
\begin{tabular}{c|c|c|c|c} 
\toprule
\multirow{2}{*}{\textbf{Dataset}} & \multirow{2}{*}{\textbf{Language}} & \multicolumn{2}{c|}{\textbf{Size of Test Set}} & \multirow{2}{*}{\textbf{\textbf{License}}}  \\ 
\cline{3-4}
                                  &                                    & \textbf{\# Queries} & \textbf{\# Document}       &                                             \\ 
\hline
NFCorpus                          & English                            & 323                 & 3K                     & CC-BY-SA-4.0                                            \\ 
\hline
SciFact                           & English                            & 300                 & 5K                     & CC-BY-NC-2.0                                \\ 
\hline
ArguAna                           & English                            & 1K                & 8K                     & CC-BY-SA-4.0                                \\ 
\hline
SCIDOCS                       & English                            & 1K                  & 25K                    & CC-BY-4.0                                   \\ 
\hline
Climate-FEVER                      &    English                                &      1K	               &   5M                       & CC-BY-SA-4.0                                             \\ 
\hline
TREC-COVID                       &  English                                  &        50	             &   171K                       & CC-BY-SA-4.0                                              \\ 
\hline
Quora                            &     English                               &         10K	            &   523K                       &  CC-BY-SA-4.0                                           \\ 
\hline
GermanQuAD                         & German                                   &              2K	       &    2M                      &  CC-BY-4.0                                           \\ 
\hline
ViHealthQA                & Vietnamese                         & 2K                  & 9K                     & -                                           \\
\hline
MA-Amazon                & English                         &  8K                 & 164K                     & Apache-2.0                                           \\
\bottomrule
\end{tabular}}
\label{tab:dataset}
\end{table}

\begin{table}
\centering
\caption{Details of models used. Some models did not clearly report the licenses in the paper or a repository.}
\scalebox{0.62}{
\begin{tabular}{c|c|c|c} 
\toprule
\textbf{Model}  & \textbf{Language} & \begin{tabular}[c]{@{}c@{}}\textbf{Number of }\\\textbf{Parameters}\end{tabular} & \textbf{License}  \\ 
\hline
\begin{tabular}[c]{@{}c@{}}PEGASUS\\(Summarization)\end{tabular}         & English           & 569M                                                                             & Apache-2.0        \\ 
\hline
\begin{tabular}[c]{@{}c@{}}T5-Base\\(QGen)\end{tabular}     & Multilingual           & 109M                                                                             & Apache-2.0        \\ 
\hline
\begin{tabular}[c]{@{}c@{}}mT5-Base\\(QGen)\end{tabular}       & Multilingual      & 390M                                                                             & Apache-2.0        \\ 
\hline
\begin{tabular}[c]{@{}c@{}}Flan T5-Base\\(Flan)\end{tabular}     & Multilingual           & 248M                                                                             & Apache-2.0        \\ 
\hline
OpenLLaMA       & Multilingual      & 3B                                                                               & Apache-2.0        \\ 
\hline
M-Distilled USE & Multilingual      & 135M                                                                             & Apache-2.0        \\ 
\hline
All-MPNet       & English           & 109M                                                                             & Apache-2.0        \\ 
\hline
Distilled-BERT  & English           & 66M                                                                              & Apache-2.0        \\ 
\hline
SGPT            & English           & 125M                                                                             & MIT               \\ 
\hline
V-SBERT         & Vietnamese        & 135M                                                                             & -                 \\ 
\hline
V-SimeCSE       & Vietnamese        & 135M                                                                             & -                 \\ 
\hline
G-Electra       & German            & 110M                                                                             & -                 \\ 
\hline
G-XLM-R         & German            & 278M                                                                             & MIT               \\ 
\hline
G-BERT          & German            & 109M                                                                             & MIT               \\ 
\hline
TAS-B           & English           & 66M                                                                              & Apache-2.0        \\ 
\hline
Contriever      & English           & 109M                                                                             & CC-BY-NC-4.0      \\ 
\hline
SPLADE++        & English           & 139M                                                                             & Apache-2.0        \\ 
\hline
ANCE            & English           & 124M                                                                             & Apache-2.0        \\ 
\hline
COCO-DR         & English           & 109M                                                                             & MIT               \\ 
\hline
DRAGON+         & English           & 109M                                                                             & CC-BY-NC-4.0      \\

\hline
BIBERT         & English           & \textasciitilde{}109M                                                                             & -      \\

\hline
MTBERT         & English           & \textasciitilde{}109M                                                                             & -      \\

\hline
MADRAL         & English           & \textasciitilde{}109M                                                                             & -      \\

\hline
ATTEMPT         & English           & \textasciitilde{}109M                                                                             & Apache-2.0      \\

\bottomrule
\end{tabular}}
\label{tab:model}
\end{table}

\noindent
\textbf{UDL Details}\;\; 
For the UDL, we tested three different methods (Concatenation, Summarization, Random permutation of the order) to link the two closest documents where we empirically selected Concatenation at last (Table \ref{tab:marge_approach_appendix}). We generated three synthetic queries for each linked and unlinked documents, noting that there is a limitation to improvements based on size (Table \ref{tab:size_query_appendix}). To decide the similarity model, we considered scikit-learn \footnote{\url{https://scikit-learn.org/stable/}} for TF-IDF, while All-MPNet \cite{song2020mpnet}, V-SBERT \cite{phobert}, and G-BERT \cite{chan-etal-2020-germans} were used for English, Vietnamese, and German datasets in pre-trained LM. The spaCy \cite{Honnibal_spaCy_Industrial-strength_Natural_2020} is utilized to import the $N_{g}$ ($en\_core\_web\_trf$ \footnote{\url{https://spacy.io/models/en}}) and $N_{s}$ ($en\_core\_sci\_scibert$ \footnote{\url{https://allenai.github.io/scispacy/}}). As shown in Tables \ref{tab:similarity_model_appendix} and \ref{tab:threshold_appendix}, we empirically decided the hyperparameters ($\gamma$=0.7, $\delta$=0.4) to get the promising results. For datasets with more than 1M documents, we considered a maximum 30K documents during query augmentations and UDL to meet the resource constraints, except for MA-Amazon where we used 60K documents. We trained the retrieval model three times with different random seeds to account for random initialization. Currently, our suggested algorithm, UDL, will follow the MIT license.

\begin{table}
\centering
\caption{Details of NER models used.}
\scalebox{0.565}{
\begin{tabular}{c|l|l} 
\toprule
                                                            & \multicolumn{1}{c|}{\textbf{\boldmath{General NER ($N_{g}$)}}}                                                                                                                                                      & \multicolumn{1}{c}{\textbf{\boldmath{Specialized NER ($N_{s}$)}}}                                                                                                                                                                                             \\ 
\hline
\begin{tabular}[c]{@{}c@{}} \textbf{Types of}\\\textbf{Entities}\end{tabular} & \begin{tabular}[c]{@{}l@{}}\textit{General:} Numerals,~Date,\\Event, Objects,~Countries,\\Language, ~Person, Quantity\\Monetary, Time, Companies,~\\Mountain ranges ...\end{tabular} & \begin{tabular}[c]{@{}l@{}}\textit{Medical:~}Organism,\\Gene,~ Chemical,\\Pathological formation,\\Cell,~Tissue ...\\\textit{Scientific:}~Task, Method,\\Metric, Material, Professional\\and Generic terms ...\end{tabular}  \\ 
\hline
\textbf{Sources}                                                     & \begin{tabular}[c]{@{}l@{}}OntoNotes 5 \cite{OntoNotes}\\ClearNLP \cite{ClearNLP}\\WordNet 3.0 \cite{Fellbaum2005-FELWAW}\\RoBERTa-Base \cite{liu2019roberta}\end{tabular}                                                                                                & \begin{tabular}[c]{@{}l@{}}OntoNotes 5 \cite{OntoNotes}\\Common Crawl \cite{common_crawl}\\GENIA 1.0 \cite{genia}\\SciBERT \cite{beltagy2019scibert}\end{tabular}                                                                                                                             \\ 
\hline
\begin{tabular}[c]{@{}c@{}}\textbf{Vocabulary}\\\textbf{Size}\end{tabular}    & 50K                                                                                                                                                                                       & 785K                                                                                                                                                                                                                            \\ 
\hline
\textbf{License}                                                     & MIT                                                                                                                                                                                       & CC-BY-SA-3.0                                                                                                                                                                                                                    \\
\bottomrule
\end{tabular}
}
\label{tab:ner}
\end{table}

\section{Notes on Reproducibility}

\label{app:comp}
\noindent
\textbf{Total Computational Budget and Infrastructure used}\;\;
For UDL and fine-tuning the retrieval models, we employed the Intel(R) Xeon(R) CPU @ 2.20GHz and NVIDIA A100. All of them used RAM 80GB and we trained three times with different seeds to get the averaged results. For decision of similarity model, TF-IDF required about 34 seconds and LM needed about 174 seconds for 10K documents. For decision of similarity score, it took about 787 seconds for 10K documents. The query augmentation for 10K documents took about 6699 seconds for summarization, 2970 seconds for Flan, 12542 seconds for OpenLLaMA and 721 seconds for QGen. Other augmentations like random cropping and RM3 are fast enough to be negligible. Fine-tuning is affected heavily by the size of the model and synthetic queries. For example, it took about 20 seconds when training a 135M parameters model with 11K queries and 4K documents. Note that, these computational costs do not affect the inference time during retrieval. In all experiments, we mainly utilized the BEIR environment \cite{thakur2021beir, kamalloo2023resources} to evaluate the retrieval performances. 

\noindent
\textbf{Hyperparameters}\;\;
In Table \ref{tab:hyperparameter}, we cover all the hyperparameters considered in this work which are based on the empirical results. During fine-tuning, we used \textit{MultipleNegativesRankingLoss} \footnote{\url{https://www.sbert.net/docs/package_reference/losses.html}} with \textit{AdamW (warmup scheduler=10\% of train set)} \cite{loshchilov2019decoupled}. During the evaluation, \textit{cosine-similarity} is utilized to retrieve the documents given queries. 

\begin{table}
\centering
\caption{Hyperparameters in UDL.}
\scalebox{0.75}{
\begin{tabular}{c|c} 
\toprule
\textbf{Parameter}                                              & \textbf{Setting}  \\ 
\hline
$\gamma$                                                        & 0.7                \\ 
\hline
$\delta$                                                        & 0.4               \\ 
\hline
\begin{tabular}[c]{@{}c@{}}Max features\\in TF-IDF\end{tabular} & 36000             \\ 
\hline
Epoch                                                           & 1                 \\ 
\hline
Learning Rate                                                   & 2e-5             \\ 
\hline
Weight Decay                                                   & 1e-2              \\
\bottomrule
\end{tabular}
}
\label{tab:hyperparameter}
\end{table}

\begin{table}
\centering
\caption{The effect of size of synthetic queries generated from QGen. Retrieval model is Distilled-BERT.}
\scalebox{0.71}{
\begin{tabular}{c|c|c|c} 
\toprule
\multirow{2}{*}{\begin{tabular}[c]{@{}c@{}}\textbf{}\\\textbf{Metrics}\end{tabular}} & \multicolumn{3}{c}{\textbf{NFCorpus}}                                                                                                                                                                                                                                                                                                                                  \\ 
\cline{2-4}
                                                                                     & \begin{tabular}[c]{@{}c@{}}{1 synthetic}\\{{queries}}\end{tabular} & \begin{tabular}[c]{@{}c@{}}{3 synthetic{{{}}}}\\{{{{queries}}}}\end{tabular} & \begin{tabular}[c]{@{}c@{}}{9 synthetic{{{}}}}\\{{{{queries}}}}\end{tabular}  \\ 
\hline
N@1                                                                               & 35.9                                                                                 & \textbf{36.9}                                                                                                                     & 36.2                                                                                                                               \\ 
\hline
N@10                                                                              & 27.9                                                                                 & \textbf{29.0}                                                                                                                     & 28.4                                                                                                                               \\ 
\hline
N@100                                                                             & 25.0                                                                                & 25.8                                                                                                                              & \textbf{26.1}                                                                                                                      \\ 
\hline
R@1                                                                             & 4.3                                                                                  & \textbf{4.5}                                                                                                                       & 4.3                                                                                                                                \\ 
\hline
R@10                                                                            & 13.2                                                                                 & \textbf{13.6}                                                                                                                     & 13.4                                                                                                                               \\ 
\hline
R@100                                                                           & 26.0                                                                                & \textbf{27.1}                                                                                                                             & 26.3                                                                                                                      \\ 
% \hline
% \textbf{Avergage}                                                                         & 22.05                                                                                 & \textbf{22.55}                                                                                                                     & 22.46                                                                                                                               \\
\bottomrule
\end{tabular}}
\label{tab:size_query_appendix}
\end{table}

\section{Term Entropy in UDL}
Equation (\ref{eq:entropy}) explains the term entropy measurement used in UDL. 

\begin{equation}
    \begin{array}{l}
    E (X)\ = -\sum_{i=1}^{N}\;P(X_{i})\log _{2} P(X_{i})
    \end{array}
\label{eq:entropy}
\end{equation}

where $E$ is the entropy, $X$ is the term, $P(X_{i})$ is the distribution of terms across documents, $N$ is the number of documents.

\begin{table}
\centering
\caption{Examples of terms from TF-IDF according to the Shannon Entropy.}
\scalebox{0.8}{
\begin{tabular}{l|l} 
\toprule
\textbf{Shannon Entropy} & \textbf{Examples of Terms}                                                                                                                                               \\ 
\hline
Greater than 1           & \begin{tabular}[c]{@{}l@{}}the, this, an, a, yes, no, is, \\was, has, have, old, new ...\end{tabular}                                                             \\ 
\hline
Less than 1              & \begin{tabular}[c]{@{}l@{}}hala, storms, ipad, sari, coax, \\intermediate, pulse,\\peculiarities, swearing, \\enlisting, endures, fervour ...\end{tabular}  \\
\bottomrule
\end{tabular}
}
\label{tab:entropy-term}
\end{table}

\begin{table}
\centering
\caption{Decisions of similarity model and type of document from UDL in each dataset.}
\scalebox{0.75}{
\begin{tabular}{c|c|c} 
\toprule
\multirow{2}{*}{\textbf{Dataset}} & \multicolumn{2}{c}{\textbf{Decisions of the UDL}}                                \\ 
\cline{2-3}
                                  & \textbf{Model} & \multicolumn{1}{l}{\textbf{Type of Document}}  \\ 
\hline
NFCorpus                          & LM             & Specialized                                     \\ 
\hline
SciFact                           & TF-IDF         & Specialized                                     \\ 
\hline
ArguAna                           & LM             & General                                        \\ 
\hline
SCIDOCS                           & LM             & Specialized                                     \\ 
\hline
Climate-FEVER                     & TF-IDF         & General                                        \\ 
\hline
TREC-COVID                        & TF-IDF         & Specialized                                     \\ 
\hline
Quora                             & LM             & General                                        \\ 
\hline
GermanQuAD                         & TF-IDF         & General                                        \\ 
\hline
ViHealthQA                        & LM             & Specialized                                     \\
\hline
MA-Amazon                        & LM             & General                                     \\
\bottomrule
\end{tabular}
}
\label{tab:dist_udl}
\end{table}

\section{Ablation Study}
\label{app:ablation}
\noindent
\textbf{Detailed Investigation of UDL}\;\;
Table \ref{tab:size_query_appendix} shows the limitation of improvement after increasing the size of synthetic queries which confirms the importance of UDL. Table \ref{tab:entropy-term} shows the examples of term entropy where article and relatively common words have entropy greater than 1 while the professional and relatively uncommon words have entropy less than 1. Table \ref{tab:dist_udl} covers the overall decisions of UDL in each dataset. 
% Figure \ref{fig:rank_dist} reveals the distribution of rank of correctly classified test queries where the distributions with UDL are more compact, have fewer outliers and allocate higher ranks for relevant document. 
Tables \ref{tab:similarity_model_appendix} and \ref{tab:threshold_appendix} reveal the details of ablation studies for hyperparameters in UDL. Table \ref{tab:marge_approach_appendix} explains the results depending on the different merging methods in UDL. Compared with random permutation, concatenation gives better results which reveals the importance of the order of sentences. Compared with summarization, concatenation shows better results which confirms the importance of the original structure of sentences.

\begin{algorithm}[H]
 \KwData{Train queries and documents in each dataset and synthetic queries}
\KwResult{Sufficient quality of synthetic queries to map the used documents}
 \textbf{Parameters:} 
 Queries in train set $Q$ = \{$q_1$ ... $q_n$\}, synthetic queries $\hat{Q}$ = \{$\hat{q}{_1}$ ... $\hat{q}_{m}$\}, documents used for generating synthetic queries and mapped by train queries $Doc$ = \{${doc_1}$ ... $doc_k$\} 

\;\>{1.}\;Find train queries mapping the linked documents in UDL:
   $q_{i}$, $doc_{a}$, $doc_{b}$\\[2pt]    
 2.\;Measure cosine-similarity in pairs of $q_{i}$-$doc_{a}$, $q_{i}$-$doc_{b}$: 
    $Score (q_{i}, doc_{a})$, $Score (q_{i}, doc_{b})$ \\[2pt]
 3.\;Measure cosine-similarity in pairs of $\hat{q}_{j}$-$doc_{a}$, $\hat{q}_{j}$-$doc_{b}$ where $\hat{q}_{j}$ is generated from linked $doc_{a}$-$doc_{b}$:
    $Score (\hat{q}_{j}, doc_{a})$, $Score (\hat{q}_{j}, doc_{b})$ \\[2pt]
\;\>{4.}\;\eIf{$Score (q_{i}, doc_{a})$ < $Score (\hat{q_{j}}, doc_{a})$ \& \\
         $Score (q_{i}, doc_{b})$ < $Score (\hat{q}_{j}, doc_{b})$}{
    $\hat{q_{j}}$\;properly maps both documents}
    {
    \If {$Score (q_{i}, doc_{a})$ < $Score (\hat{q}_{j}, doc_{a})$}
    {
    $\hat{q_{j}}$\;appropriately maps $doc_{a}$}
    
    \If {$Score (q_{i}, doc_{b})$ < $Score (\hat{q}_{j}, doc_{b})$}
    {
    $\hat{q}_{j}$\;appropriately maps $doc_{b}$}
}
\caption{Quality Checking}
\label{algo:quality}
\end{algorithm}

\noindent
\textbf{Quality of Synthetic Queries}\;\;
Algorithm \ref{algo:quality} reveals the overall logic of quality checking based on the offered train set in NFCorpus and SciFact. We first found train data which covers same documents considered as linking in UDL. Then, we measured the cosine-similarity between the train query and relevant documents, and compared this with the cosine-similarity between the generated synthetic query and those same documents. If generated query has higher scores, this argues that our generated data has enough quality to link the single/multiple documents. 

From our analysis, 93\% of generated queries properly maps both documents where it increases up to 99\% for single document. Thus, most of queries generated from linked documents in UDL have the sufficient quality to map the relevant documents without additional quality control.

% This also shows hyperparameters-tuning in UDL is enough to control the quality of synthetic queries without additional quality control. 

%%%%%%%%%%%%%%%%%%%%%%%
\begin{table*}
\centering
\caption{Different similarity models for UDL. Retrieval model is Distilled-BERT and similarity score is 0.6 for NFCorpus, Scifact and 0.4 for ArguAna. $\gamma=0.7$ is our final decision.}
\scalebox{0.65}{
\begin{tabular}{c|c|c|c|c|c|c|c|c|c|c|c|c} 
\toprule
\multirow{2}{*}{\textbf{Metrics}} & \multicolumn{4}{c|}{\textbf{NFCorpus}}                     & \multicolumn{4}{c|}{\textbf{\textbf{SciFact}}}              & \multicolumn{4}{c}{\textbf{ArguAna}}                         \\ 
\cline{2-13}
                                  & $\gamma=0.1$ & $\gamma=0.3$ & \boldmath{$\gamma=0.7$}  & $\gamma=0.9$ & $\gamma=0.1$ & $\gamma=0.3$  & \boldmath{$\gamma=0.7$}  & $\gamma=0.9$ & $\gamma=0.1$ & $\gamma=0.3$  & \boldmath{$\gamma=0.7$}  & $\gamma=0.9$  \\ 
\hline
N@1                            & 37.7         & 37.6         & \textbf{39.0} & 35.8         & 49.2         & 49.0          & \textbf{50.4} & 49.6         & 29.2         & 30.1          & \textbf{30.3} & 27.7          \\ 
\hline
N@10                           & 30.5         & 30.4         & \textbf{31.2} & 28.9         & 60.1         & 60.1          & \textbf{61.5} & 61.1         & 54.6         & 55.2          & \textbf{55.8} & 53.9          \\ 
\hline
N@100                          & 28.4         & 28.5         & \textbf{28.9} & 25.2         & 65.1         & \textbf{65.2} & 64.9          & 64.1         & 57.9         & \textbf{59.2} & \textbf{59.2} & 55.4          \\ 
\hline
R@1                          & 4.3          & 4.3          & \textbf{4.4}  & 3.9          & 46.8         & 46.5          & \textbf{48.1} & 48.0         & 29.0           & 29.5          & \textbf{30.3} & 27.7          \\ 
\hline
R@10                         & 14.2         & 14.3         & \textbf{14.7} & 13.2         & 75.2         & 72.5          & \textbf{73.3} & 73.2         & 84.0         & 84.3          & \textbf{85.1} & 78.8          \\ 
\hline
R@100                        & 30.1         & 30.3         & \textbf{30.8} & 27.8         & 88.4         & 88.2          & \textbf{90.7} & 90.2         & 99.1         & 98.7          & \textbf{99.2} & 98.4          \\
\bottomrule
\end{tabular}}
\label{tab:similarity_model_appendix}
\end{table*}
\begin{table*}
\centering
\caption{Different similarity scores for UDL. Retrieval model is Distilled-BERT and similarity model is fixed to TF-IDF. $\delta = 0.4$ is our final choice.}

\scalebox{0.65}{
\begin{tabular}{c|c|c|c|c|c|c|c|c|c|c|c|c} 
\toprule
\multirow{2}{*}{\textbf{Metrics}} & \multicolumn{4}{c|}{\textbf{NFCorpus}}                                        & \multicolumn{4}{c|}{\textbf{\textbf{SciFact}}}                                & \multicolumn{4}{c}{\textbf{ArguAna}}                        \\ 
\cline{2-13}
                                  & $\delta=0.2$ & \boldmath{$\delta=0.4$}           & $\delta=0.6$           & $\delta=0.8$ & $\delta=0.2$ & \boldmath{$\delta=0.4$}           & $\delta=0.6$           & $\delta=0.8$ & $\delta=0.2$ & \boldmath{$\delta=0.4$}  & $\delta=0.6$ & $\delta=0.8$  \\ 
\hline
N@1                            & 37.4         & \textbf{\textbf{39.2}} & 36.7                   & 37.2         & 44.0         & \textbf{\textbf{50.4}} & 47.3                   & 47.3         & 25.6         & \textbf{26.8} & 25.6         & 25.4          \\ 
\hline
N@10                           & 28.1         & \textbf{\textbf{29.0}} & 28.6                   & 28.1         & 57.9         & \textbf{\textbf{61.5}} & 59.3                   & 58.8         & 50.9         & \textbf{51.5} & 50.3         & 49.5          \\ 
\hline
N@100                          & 25.3         & \textbf{\textbf{26.3}} & 26.1                   & 26.0         & 60.8         & \textbf{\textbf{64.9}} & 63.2                   & 62.6         & 54.6         & \textbf{55.7} & 54.6         & 54.1          \\ 
\hline
R@1                          & 4.4          & \textbf{\textbf{4.6}}  & 3.8                    & 4.0          & 41.8         & \textbf{\textbf{48.1}} & 44.9                   & 44.8         & 25.6         & \textbf{26.8} & 25.6         & 25.1          \\ 
\hline
R@10                         & 12.8         & 12.9                   & \textbf{\textbf{13.4}} & 13.2         & 71.2         & 73.3                   & \textbf{\textbf{73.9}} & 71.4         & 79.3         & \textbf{80.1} & 79.3         & 77.0          \\ 
\hline
R@100                        & 25.9         & \textbf{\textbf{27.3}} & 26.6                   & 26.1         & 88.3         & \textbf{\textbf{90.7}} & 89.6                   & 90.1         & 97.4         & \textbf{98.4} & 97.9         & 97.2          \\
\bottomrule
\end{tabular}
}

\label{tab:threshold_appendix}
\end{table*}

\begin{table*}
\centering
\caption{Results according to the merging approaches in UDL. Random permutation: Concatenate two documents and then, randomly mix up the order. Summarization: Using Flan T5-Base \cite{flan}, summarize each document separately and then, concatenate them. Title is always attached directly.}
\scalebox{0.61}{\begin{tabular}{c|c|c|c|c|c|c|c|c|c} 
\toprule
\multirow{3}{*}{\textbf{Metrics}} & \multicolumn{3}{c|}{\textbf{NFCorpus}}                                                         & \multicolumn{3}{c|}{\textbf{SciFact}}                                                          & \multicolumn{3}{c}{\textbf{ArguAna}}                                                            \\ 
\cline{2-10}
                                  & \textbf{Concatenation}    & \begin{tabular}[c]{@{}c@{}}Random\\Permutation\end{tabular} & Summarization & \textbf{Concatenation}    & \begin{tabular}[c]{@{}c@{}}Random\\Permutation\end{tabular} & Summarization & \textbf{Concatenation}    & \begin{tabular}[c]{@{}c@{}}Random\\Permutation\end{tabular} & Summarization  \\ 
\hline
N@1                            & \textbf{39.0} & 37.5                                                     & 38.6       & \textbf{50.4} & 47.3                                                     & 48.3       & \textbf{30.3} & 29.6                                                     & 23.4        \\ 
\hline
N@10                           & \textbf{31.2} & 30.0                                                     & 29.6       & \textbf{61.5} & 58.9                                                     & 59.4       & \textbf{55.8} & 54.8                                                     & 45.9        \\ 
\hline
N@100                          & \textbf{28.9} & 28.4                                                     & 28.0       & \textbf{64.9} & 62.6                                                     & 63.4       & \textbf{59.2} & 58.1                                                     & 51.5        \\ 
\hline
R@1                          & \textbf{4.4}  & 4.0                                                      & 4.3        & \textbf{48.1} & 44.9                                                   & 45.9       & \textbf{30.3} & 30.0                                                     & 23.4        \\ 
\hline
R@10                         & \textbf{14.7} & 14.2                                                     & 13.5       & \textbf{73.3} & 72.5                                                     & 72.0       & \textbf{85.1} & 83.9                                                     & 73.7        \\ 
\hline
R@100                        & \textbf{30.8} & 30.1                                                     & 30.0       & \textbf{90.7} & 89.2                                                     & 90.3       & \textbf{99.2} & 98.7                                                     & 98.0        \\ 
% \hline
% \textbf{Avg}                      & \textbf{24.82} & 24.03                                                     & 24          & \textbf{64.75} & 62.56                                                     & 63.22       & \textbf{59.97} & 59.68                                                     & 52.64        \\
\bottomrule
\end{tabular}}
\label{tab:marge_approach_appendix}
\end{table*}